\documentclass[10pt, conference, compsocconf]{IEEEtran}
\IEEEoverridecommandlockouts

\usepackage[utf8]{inputenc} 
\usepackage[T1]{fontenc}    
\usepackage[sort&compress, numbers]{natbib}
\usepackage{booktabs}       
\usepackage{nicefrac}       
\usepackage{microtype}      
\usepackage{graphicx}
\usepackage{caption}
\usepackage{subcaption}
\usepackage{amsfonts,mathtools}
\usepackage{algorithmic}
\usepackage[]{algorithm}
\usepackage{array}
\usepackage{bm}
\usepackage{mathrsfs}
\usepackage{xcolor}
\usepackage{float}
\usepackage{epstopdf}
\usepackage{amssymb}
\usepackage{amsmath}
\usepackage{multirow}
\usepackage{hyperref} 

\usepackage[capitalize,noabbrev]{cleveref}

\renewcommand{\baselinestretch}{1.05}

\usepackage{amsthm} 

\renewcommand{\algorithmicrequire}{ \textbf{Input:}} 
\renewcommand{\algorithmicensure}{ \textbf{Output:}}

\mathchardef\mhyphen="2D

\makeatletter
\newcommand*\rel@kern[1]{\kern#1\dimexpr\macc@kerna}

\newcommand*\widebar[1]{%
  \kern 0.18em
  \hbox{%
    \vbox{%
      \hrule height 0.5pt 
      \kern0.35ex
      \hbox{%
        \kern-0.18em
        \ensuremath{#1}%
        \kern-0.18em
      }%
    }%
  }%
  \kern 0.18em
} 

\theoremstyle{plain}
\newtheorem{theorem}{Theorem}

\theoremstyle{definition}
\newtheorem{definition}[theorem]{Definition}

\theoremstyle{remark}

\newcommand{\norm}[1]{\ensuremath{\left\|#1\right\|}}	
\providecommand{\tr}[1]{\text{tr}\left(#1\right)}
\providecommand{\norm}[1]{\left \| #1 \right \|}

\newcommand{\normI}[1]{{\left\vert\kern-0.25ex\left\vert\kern-0.25ex\left\vert #1 
    \right\vert\kern-0.25ex\right\vert\kern-0.25ex\right\vert}}

\makeatletter
\newcommand{\linebreakand}{%
  \end{@IEEEauthorhalign}
  \hfill\mbox{}\par
  \mbox{}\hfill\begin{@IEEEauthorhalign}
}
\makeatother

\begin{document}

\title{Network Topology Inference with Sparsity and Laplacian Constraints}

\author{\IEEEauthorblockN{Jiaxi Ying}
\IEEEauthorblockA{Hong Kong University of Science and Technology\\
Email: jx.ying@connect.ust.hk}
\and
\IEEEauthorblockN{Xi Han$^*$}
\IEEEauthorblockA{North China University of Technology\\
Email: hanxi@ncut.edu.cn \\
* Corresponding author}
\linebreakand
\IEEEauthorblockN{Rui Zhou}
\IEEEauthorblockA{Shenzhen Research Institute of Big Data\, \\
Email: rui.zhou@sribd.cn} 
\and
\IEEEauthorblockN{Xiwen Wang}
\IEEEauthorblockA{Hong Kong University of Science and Technology\, \\
Email: xwangew@connect.ust.hk}
\and
\IEEEauthorblockN{Hing Cheung So}
\IEEEauthorblockA{City University of Hong Kong\\
Email: hcso@ee.cityu.edu.hk \thanks{This work was supported in part by the National Nature Science Foundation of China (NSFC) under Grant 62001008 and 62201362, in part by the Shenzhen Science and Technology Program (Grant No. RCBS20221008093126071), and in part by Beijing Natural Science Foundation 4212002.}}
}

\maketitle

\begin{abstract}
We tackle the network topology inference problem by utilizing Laplacian constrained Gaussian graphical models, which recast the task as estimating a precision matrix in the form of a graph Laplacian. Recent research \cite{ying2020nonconvex} has uncovered the limitations of the widely used $\ell_1$-norm in learning sparse graphs under this model: empirically, the number of nonzero entries in the solution grows with the regularization parameter of the $\ell_1$-norm; theoretically, a large regularization parameter leads to a fully connected (densest) graph. To overcome these challenges, we propose a graph Laplacian estimation method incorporating the $\ell_0$-norm constraint. An efficient gradient projection algorithm is developed to solve the resulting optimization problem, characterized by sparsity and Laplacian constraints. Through numerical experiments with synthetic and financial time-series datasets, we demonstrate the effectiveness of the proposed method in network topology inference.
\end{abstract}


\IEEEpeerreviewmaketitle

\section{Introduction}\label{sec-1}

In modern signal processing applications, the analysis of signals residing on networks, often referred to as graph signals, has become increasingly important \cite{shuman2013emerging,ortega2018graph,dong2019learning,mateos2019connecting,9244648,leus2023graph}. These graph signals emerge in a variety of fields, such as data gathered from wireless sensor networks and electroencephalography (EEG) signals recorded in brain connectivity networks. Laplacian constrained Gaussian graphical models (GGMs) \cite{ying2020does,kumar2019unified} provide a powerful tool for characterizing these signals on smooth graphs \cite{dong2016learning}, where a substantial edge weight between two vertices indicates a high similarity in their signal values. In this paper, we tackle the network topology inference problem under Laplacian constrained GGMs, which recasts the task as estimating the precision matrix (i.e., inverse covariance matrix) as a graph Laplacian in a multivariate Gaussian distribution. The zero pattern of the precision matrix reveals the network topology, offering valuable insights into how these sensors interact, which can be used to optimize the network's performance and reliability.

GGMs have been widely explored in the literature, with the graphical lasso \cite{banerjee2008model, d2008first} serving as a prominent estimation method. This approach leverages an $\ell_1$-norm regularized Gaussian maximum likelihood estimation, which has proven effective in imposing sparsity on the solution. With larger $\ell_1$-norm regularization parameters, the solution becomes increasingly sparse. In this paper, our focus lies on Laplacian constrained GGMs, wherein the precision matrix takes the form of a graph Laplacian. Interestingly, recent studies \cite{ying2020nonconvex,ying2020does} have revealed that applying the $\ell_1$-norm to learn Laplacian constrained GGMs results in an increased number of nonzero entries as the regularization parameter grows, yielding dense graphs rather than sparse ones. While nonconvex regularization overcomes this issue \cite{ying2020nonconvex,ying2020does}, it necessitates tuning multiple parameters. We introduce a graph Laplacian estimation method incorporating the $\ell_0$-norm constraint, which is the most intuitive and natural approach to control the sparsity of solutions.

Laplacian constrained GGMs have attracted growing interest in the fields of signal processing and machine learning over graphs \cite{ying2021minimax,egilmez2017graph,zhao2019optimization}. Recent work \cite{ying2021minimax} has established that the maximum likelihood estimator (MLE) under Laplacian constraints exists with as few as one observation, irrespective of the underlying dimension. This finding significantly reduces the sample size requirement from $n \geq p$ in general GGM cases, where $n$ and $p$ denote the sample size and problem dimension, respectively. Precision matrices in Laplacian constrained GGMs take the form of a graph Laplacian, which enables the interpretation of the eigenvalues and eigenvectors as spectral frequencies and Fourier bases \cite{segarra2017network}. Structured graph learning has been explored by leveraging spectral graph theory \cite{kumar2019unified,cardoso2021graphical,cardoso2022learning,kumar2019bipartite,cardoso2024nonconvex}.

Generalized Laplacian constrained GGMs have also garnered increasing attention \cite{egilmez2017graph,yuan2023joint}. These models feature nonpositive off-diagonal entries in the precision matrix, while the zero-sum condition for rows/columns is not upheld. The resulting precision matrix is a symmetric \textit{M}-matrix, and such models satisfy the total positivity property \cite{fallat2017total,lauritzen2019maximum,rossell2021dependence}, a strong form of positive dependence. The works \cite{lauritzen2019maximum,slawski2015estimation} have demonstrated that the MLE for these models exists if the sample size meets the condition $n \geq 2$. One approach to estimating a generalized graph Laplacian is the MLE \cite{lauritzen2019maximum,slawski2015estimation}, which implicitly promotes sparsity through the \textit{M}-matrix constraint. The (weighted) $\ell_1$-norm regularized MLE \cite{egilmez2017graph,ying2021fast} provides improved sparsity control, and several algorithms have been developed to tackle it, such as block coordinate descent \cite{egilmez2017graph, pavez2016generalized}, proximal point algorithm \cite{deng2020fast}, and projected Newton-like methods \cite{cai2021fast}. The estimation of diagonally dominant \textit{M}-matrices as precision matrices has been studied in \cite{egilmez2017graph,ying2023adaptive,truell2021maximum}.

In this paper, we investigate the network topology inference problem by estimating the precision matrix as a graph Laplacian under a sparsity constraint. It is important to note that conventional estimation methods for general GGMs, like graphical lasso, typically utilize sparsity-promoting regularization to learn sparse graphs, as the $\ell_0$-constrained formulation does not yield optimal solutions when the sample size is smaller than the dimension (i.e., $n < p$). Our paper presents three main contributions:
\begin{itemize}
\item We propose a graph Laplacian estimation method that incorporates the $\ell_0$-norm constraint, addressing the shortcomings of the $\ell_1$-norm regularization when estimating Laplacian constrained GGMs. We establish that the existence of optimal solutions can be guaranteed under Laplacian constraints, even when $n = 1$.

\item We devise an efficient gradient projection algorithm to solve the resulting estimation problem with sparsity and Laplacian constraints.

\item We conduct numerical experiments on both synthetic and real-world datasets, demonstrating the effectiveness of our proposed method in inferring network topologies.
\end{itemize}

\paragraph{Notation} $\norm{\bm x}$ and $\norm{\bm x}_0$ denote Euclidean norm and the number of nonzero entries, respectively. $\mathbb{S}_+^p$ and $\mathbb{S}_{++}^p$ denote the sets of positive semi-definite and positive definite matrices with the dimensions $p \times p$, respectively. $\mathbb{R}_+^p$ represents the set of all $p$-dimensional vectors with non-negative real-valued components. $[p]$ denotes the set $\{1, \ldots, p\}$.

\section{Background and Problem Formulation}\label{sec-2}
In this section, we first provide an introduction to Laplacian constrained GGMs, and then present the problem formulation.

\subsection{Laplacian Constrained Graphical Models}\label{sec-background}

We define a weighted, undirected graph $\mathcal{G}=(V,E,\bm W)$, where $V$ denotes the set of vertices, $E$ represents the set of edges, and $\bm W \in \mathbb{R}^{p \times p}_+$ is the weighted adjacency matrix with $W_{ij}$ denoting the graph weight between vertex $i$ and vertex $j$. The graph Laplacian $\bm L$ is defined as:
\begin{equation}\label{Lap}
\bm L=\bm D-\bm W, 
\end{equation}
where $\bm D$ is a diagonal matrix where $D_{ii}=\sum_{j=1}^p W_{ij}$. Throughout this paper, we focus on connected graphs, wherein the graph comprises a single component. Following from spectral graph theory \cite{chung1997spectral}, the Laplacian matrix of a connected graph with $p$ vertices has a rank of $p - 1$. Consequently, the set of Laplacian matrices for connected graphs can be formulated as:
\begin{equation}\label{Lap-set}
\begin{split}
\mathcal{S}_L &:= \big\lbrace  \bm X \in \mathbb{S}_{+}^p \, | \, \bm X \cdot \bm 1 = \bm 0, \ \bm X = \bm X^\top,  \\
 & \qquad \qquad  X_{ij} \leq 0, \, \forall \, i \neq j, \ \mathrm{rank}(\bm X) = p-1 \big\rbrace.
\end{split}
\end{equation}
where $\bm 0$ and $\bm 1$ denote the zero and one vectors, respectively.

Let $\bm y = \left[ y_1, \ldots, y_p \right]^\top $ be a zero-mean $p$-dimensional random vector following $\bm y \sim \mathcal{N}(\bm 0, \bm \Sigma)$, where $\bm \Sigma$ is the covariance matrix. We associate the random vector $\bm y$ with a graph $\mathcal{G}=(V,E,\bm W)$. As a result, $\bm y$ forms a GGM with respect to the graph $\mathcal{G}$. When the inverse covariance matrix, also called precision matrix, is a graph Laplacian, the random vector forms a Laplacian constrained GGM. Consequently, the problem of topology inference can be transformed into graph Laplacian estimation.

\subsection{Problem Formulation}\label{sec-PF}

We consider the problem of estimating the precision matrix as a graph Laplacian, given $n$ independent and identically distributed observations $\{ \bm y^{(k)}\}_{k=1}^n$. The maximum likelihood estimation can be formulated as the following Laplacian constrained log-determinant program:
\begin{equation}\label{MLE}
\begin{array}{ll}
\underset{\bm X }{\mathsf{minimize}} & - \log \det^\star (\bm X) + \tr{\bm S \bm X},  \\  
\mathsf{subject~to} &  \bm X \in \mathcal{S}_L,
\end{array}
\end{equation}
where ${\det}^\star$ denotes the pseudo determinant defined by the product of nonzero eigenvalues \citep{holbrook2018differentiating}. It has been demonstrated in \cite{ying2020nonconvex} that the Laplacian set $\mathcal{S}_{L}$ defined in \eqref{Lap-set} can be equivalently written as 
\begin{equation}\label{Lap-set-new}
\mathcal{S}_L  = \left\lbrace \bm X \in \mathbb{R}^{p \times p} \, | \, ( \bm X + \bm J) \in \mathbb{S}^{p}_{++}, \ \bm X \in \mathcal{S}_Z,  \right\rbrace,
\end{equation}
where $\bm J=\frac{1}{p} \bm 1_{p \times p}$ is a constant matrix with each element equal to $\frac{1}{p}$, and the set $\mathcal{S}_Z$ is defined as:
\begin{equation*}
\mathcal{S}_Z := \! \left\lbrace \bm X\in \mathbb{R}^{p \times p} | \bm X \cdot \bm 1 = \bm 0, \bm X = \bm X^\top, X_{ij} \leq 0, \forall \, i \neq j  \right\rbrace.
\end{equation*}
Then Problem~\eqref{MLE} can be equivalently reformulated as:
\begin{equation}\label{cost-theta}
\begin{array}{ll}
\underset{\bm X}{\mathsf{minimize}} &  - \log \det (\bm X + \bm J) + \tr{\bm S \bm X},  \\  
\mathsf{subject~to} &  \bm X + \bm J \in \mathbb{S}_{++}^p, \, \bm X \in \mathcal{S}_Z,
\end{array}
\end{equation}
where $\bm S$ is the sample covariance matrix, constructed as $\bm S = \frac{1}{n} \sum_{k=1}^n \bm y^{(k)} \bm (\bm y^{(k)})^\top$. It is worth noting that we replace ${\det}^{\star} ( \bm X )$ with $ {\det} ( \bm X + \bm J)$ in \eqref{cost-theta}, as done in \cite{egilmez2017graph}, because the function ${\det}^{\star}$ is not continuous, which poses difficulty in developing algorithms.

\section{Proposed Method}\label{sec-3}
In this section, a new formulation incorporating sparsity and Laplacian constraints is introduced. We develop a gradient projection algorithm to solve the resulting problem.

\subsection{Sparsity Constrained Formulation}\label{sec-3-1}

The $\ell_1$-norm regularization in graphical lasso has been widely acknowledged for its effectiveness across various fields. However, the $\ell_1$-norm has been shown to be less effective in Laplacian constrained GGMs \cite{ying2020nonconvex,ying2020does}. Consequently, we propose a novel formulation incorporating the sparsity constraint to address this limitation.

We introduce a sparsity-constrained maximum likelihood estimation for Laplacian constrained GGMs:
\begin{equation}\label{MLE-theta}
\begin{array}{ll}
\underset{\bm X}{\mathsf{minimize}} & \!\!\! - \log \det (\bm X + \bm J) + \tr{\bm S \bm X},  \\  
\mathsf{subject~to} & \!\!\! \bm X + \bm J \in \mathbb{S}_{++}^p, \bm X \in \mathcal{S}_Z, \| \bm X \|_{0, \mathrm{off}} \leq 2s,
\end{array}
\end{equation}
where $\| \bm X \|_{0, \mathrm{off}}$ denotes the number of nonzero off-diagonal entries in $\bm X$. The sparsity constraint $\| \bm X \|_{0, \mathrm{off}} \leq 2s$ ensures that the learned graph is sparse, with the number of edges not exceeding a predetermined value $s$. The sparsity level $s$ can be estimated in certain tasks. For instance, for learning structured graphs with $p$ vertices, a connected tree graph has $p-1$ edges, while a circular graph has $p$ edges.

We highlight that traditional estimation methods for general GGMs, such as graphical lasso, often employ sparsity-promoting regularization rather than the $\ell_0$-norm constraint. This is because the $\ell_0$-norm constrained formulation fails to provide optimal solutions when the number of samples is less than the dimension (i.e., $n < p$).

\begin{theorem} \label{Theorem-exist}
The set of global minimizers of Problem~\eqref{MLE-theta} is nonempty and compact almost surely as long as the number of observations $n \geq 1$.
\end{theorem}
Theorem~\ref{Theorem-exist} establishes that the set of optimal solutions for the $\ell_0$-norm constrained problem in \eqref{MLE-theta} is guaranteed to be nonempty almost surely, even when $n=1$. This crucial insight forms the basis of our proposed $\ell_0$-norm approach. It is worth noting that the existence of optimal solutions for Problem~\eqref{MLE-theta} is implicitly assumed in the remainder of this paper. As demonstrated in Theorem~\ref{Theorem-exist}, this assumption holds with probability one.

\subsection{Gradient Projection Algorithm}\label{sec-3-2} 
Problem~\eqref{MLE-theta} is a nonconvex optimization problem with multiple constraints, where the constraints $X_{ij} = X_{ji}$ and $\bm X \cdot \bm 1 =\bm 0$ are linear, resulting in only $p(p-1)/2$ variables in $\bm X$ being free. To address these constraints, we employ a linear operator, as defined in \cite{kumar2019unified}, which maps a vector $\bm x \in \mathbb{R}^ {p(p-1)/2}$ to a matrix $\mathcal{L} \bm x \in \mathbb{R}^{p \times p}$.
\begin{definition}
The linear operator $\mathcal{L}: \mathbb{R}^ {p(p-1)/2}\rightarrow \mathbb{R}^{p \times p}$, $\bm x \mapsto \mathcal{L} \bm x$, is defined by
\begin{equation}\label{operator}
[\mathcal{L} \bm x]_{ij} =
\begin{cases}
-x_{k}\; & \; \;i >j,\\
\hspace{.2cm}[\mathcal{L}\bm x]_{ji} \; & \; \;i<j,\\
-\sum_{j \neq i}[\mathcal{L} \bm x]_{ij}\; &\;\; i=j,
\end{cases}
\end{equation}
where $k=i-j+\frac{j-1}{2}(2p-j)$.
\end{definition}
The adjoint operator $\mathcal{L}^*$ of $\mathcal{L}$ is defined to fulfill the condition $\langle \mathcal{L}\bm x,\bm Y \rangle=\langle \bm x,\mathcal{L}^* \bm Y \rangle$, for all $\bm x \in \mathbb{R}^{p(p-1)/2}$ and $\bm Y \in \mathbb{R}^{p \times p}$. Furthermore, it is a linear operator that maps a matrix $\bm Y \in \mathbb{R}^{p \times p}$ back to a vector $\mathcal{L}^* \bm Y \in \mathbb{R}^ {p(p-1)/2}$.
\begin{definition}
The adjoint operator $\mathcal{L}^*: \mathbb{R}^{p\times p} \rightarrow \mathbb{R}^{p(p-1)/2}$, $\bm Y \mapsto \mathcal{L}^* \bm Y$, is defined by
	\begin{equation}\label{adj-operator}
	[\mathcal{L}^* \bm Y]_k=Y_{ii}-Y_{ij}-Y_{ji}+Y_{jj},
	\end{equation}
	where $i,j \in [p]$ obeying $ k=i-j+\frac{j-1}{2}(2p-j)$ and $i>j$.
\end{definition}

For notational simplicity, we represent $p(p-1)/2$ as $d$. By incorporating the linear operator $\mathcal{L}$, we can effectively simplify Problem~\eqref{MLE-theta} as:
\begin{equation}\label{MLE-x}
\begin{array}{ll}
\underset{\bm x }{\mathsf{minimize}} & - \log \det ( \mathcal{L}\bm x + \bm J) + \tr{\bm S \mathcal{L}\bm x},  \\  
\mathsf{subject~to} & \bm x \in  \Omega_s \cap \mathbb{R}^d_+ \cap \mathbb{V}_{++}^d,
\end{array}
\end{equation}
where the sets $\Omega_s$ and $\mathbb{V}_{++}^d$ are defined as:
\begin{equation*}
\Omega_s := \big\lbrace \bm x \in \mathbb{R}^d \, | \, {\| \bm x \|}_0 \leq s \big\rbrace,
\end{equation*}
and
\begin{equation*}
\mathbb{V}_{++}^d := \big\lbrace \bm x \in \mathbb{R}^d \, | \, \mathcal{L} \bm x + \bm J \in \mathbb{S}_{++}^p \big\rbrace.
\end{equation*}

Both the sets $\Omega_s$ and $\mathbb{R}_+^p$ are closed and can be addressed using a projection $\mathcal{P}_{\Omega_s \cap \mathbb{R}_+^p}$ onto their intersection with respect to the Euclidean norm. The projection $\mathcal{P}_{\Omega_s \cap \mathbb{R}_+^p} (\bm z)$ can be computed efficiently by sorting the $p$ entries of $\mathcal{P}_{\mathbb{R}_+^p}(\bm z)$ and retaining only the $s$ largest values, while setting the remaining ones to zero. Since $\mathbb{V}_{++}$ is not closed, we employ a backtracking line search method to tackle this constraint.

Let $f$ represent the objective function of Problem~\eqref{MLE-x}. A gradient projection step at $\bm x_k$ can be constructed as:
\begin{equation}
\bm x_k (\eta_k) \in \mathcal{P}_{\mathbb{R}^p_+ \cap \Omega_s} \left(\bm x_k - \eta_k \nabla f \left(\bm x_k \right) \right),
\end{equation}
where $\eta_k$ is the step size, and $\nabla f \left(\bm x_k \right)$ denotes the gradient of $f$ at $\bm x_k$:
\begin{equation}
\nabla f \left(\bm x_k \right) = \mathcal{L}^* \big( -(\mathcal{L}\bm x_k + \bm J )^{-1} + \bm S \big).
\end{equation}
Define the gradient mapping at $\bm x_k$ as follows:
\begin{equation}
G_{\frac{1}{\eta_k}} (\bm x_k) := \frac{1}{\eta_k} \left(\bm x_k - \bm x_k \left(\eta_k \right) \right).
\end{equation}
We note that, in the unconstrained case, $G_{\frac{1}{\eta_k}} (\bm x_k)$ simplifies to $\nabla f \left(\bm x_k \right)$. As such, the gradient mapping can be considered an extension of the standard gradient operation.

We determine the step size using an Armijo-like rule, ensuring global convergence for our algorithm. Specifically, we examine step sizes $\eta_k \in \sigma\left\lbrace \beta^0, \beta^1, \beta^2, \ldots \right\rbrace$, where $\sigma >0$ and $\beta \in (0, 1)$. We seek the smallest integer $m \in \mathbb{N}$ such that the iterate $\bm x_k (\eta_k) \in \mathbb{V}_{++}$ with $\eta_k = \sigma \beta^m$, and leads to a sufficient reduction of the objective function value:
\begin{equation}\label{line-cond}
f \left( \bm x_k (\eta_k) \right) \leq f\left( \bm x_k \right) - \alpha \eta_k  \big \| G_{\frac{1}{\eta_k}} (\bm x_k) \big \|^2,
\end{equation}
where $\alpha \in (0, 1)$. The backtracking line search condition \eqref{line-cond} is a variant of the Armijo rule.

To ensure $\bm x_k (\eta_k) \in \mathbb{V}_{++}$, we need to verify the positive definiteness of $\mathcal{L} \bm x_k (\eta_k) + \bm J$. This verification can be conducted during the computation of the Cholesky factorization for objective function evaluation. A summary of our algorithm is provided in Algorithm~\ref{algo}. Furthermore, our method is adaptable for estimating various structured matrices, including Hankel matrices, through the application of the Hankel linear operator \cite{ying2017hankel,ying2018vandermonde,cai2016robust}.

\begin{algorithm}
	\caption{Proposed algorithm} \label{algo} 
		\begin{algorithmic}[1]
		\STATE {\bfseries Input:} Sample covariance matrix $\bm S$, sparsity level $s$, $\sigma >0$, $\alpha \in (0, 1)$, and $\beta \in (0, 1)$; 
		 \vspace{0.1cm}		
		\FOR { $k = 0, 1, 2, \ldots$}		
       \STATE $m \gets 0$;
       \REPEAT
       \vspace{0.1cm}
       \STATE Update $\bm x_{k+1} \in \mathcal{P}_{\Omega_s \cap \mathbb{R}^p_+} \left(\bm x_k - \sigma \beta^m \nabla f \left(\bm x_k \right) \right)$,
       \vspace{0.1cm}
       \STATE $m \gets m+1$;
       \vspace{0.1cm}
       \UNTIL {$\bm x_{k+1} \in \mathbb{V}_{++}$, and       
       $$f \left( \bm x_{k+1} \right) \leq f\left( \bm x_k \right) - \alpha \sigma \beta^m  \big \| G_{\frac{1}{\sigma \beta^m}} (\bm x_k) \big \|^2;$$}
		\ENDFOR
	\end{algorithmic}
\end{algorithm}

\section{Experimental Results}\label{sec-experiment}
We conduct numerical experiments on synthetic and real-world data to verify the performance of the proposed method.

\subsection{Synthetic Data}\label{sec_synthetic}

We carry out numerical simulations to evaluate the estimation performance of the GLE-ADMM \cite{zhao2019optimization}, NGL \cite{ying2020does}, and our proposed method. The GLE-ADMM and NGL methods use the $\ell_1$-norm and MCP penalties to estimate Laplacian-constrained precision matrices, respectively. In contrast, our method employs the $\ell_0$-norm constraint.

We randomly generate an Erdos-Renyi graph, a well-known random graph model, to serve as the underlying ground-truth graph. The graph contains 100 vertices, and the graph weights associated with the edges are uniformly sampled from $U(2, 5)$. We independently draw samples $\bm y^{(1)}, \ldots, \bm y^{(n)}$ under the Laplacian-constrained GGM.

\begin{figure}[!htb]
    \centering    
        \includegraphics[scale=.39]{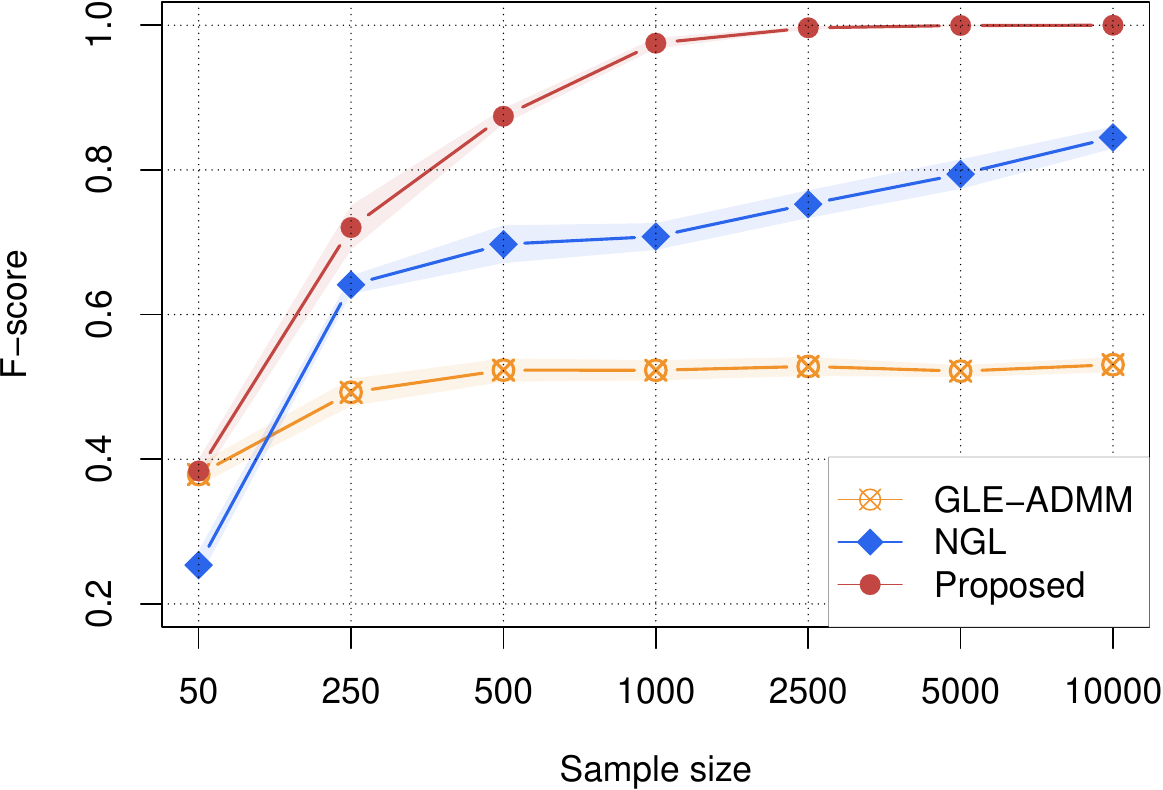}
    \caption{F-score as a function of sample size for learning Erdos-Renyi graphs consisting of 100 nodes.} \label{fig:er-fs}
\end{figure}

\begin{figure*}[!htb]
    \captionsetup[subfigure]{justification=centering}
    \centering
      \begin{subfigure}[t]{0.3\textwidth}
        \centering
        \includegraphics[scale=.36]{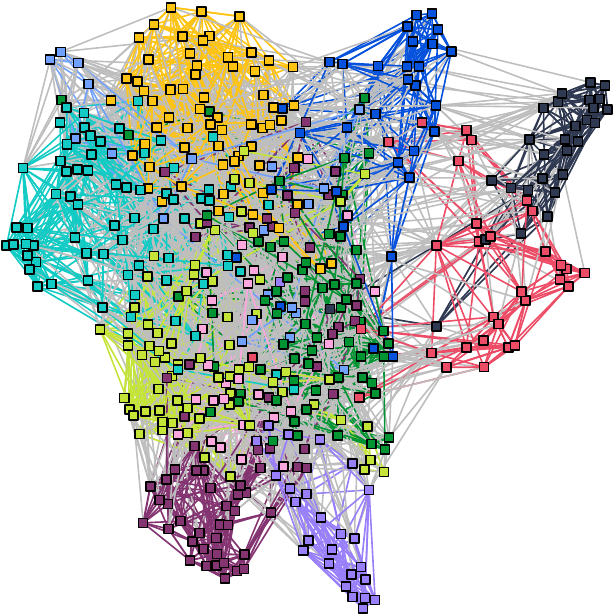}
        \caption{GLE-ADMM}
    \end{subfigure}%
        \begin{subfigure}[t]{0.3\textwidth}
        \centering
        \includegraphics[scale=.36]{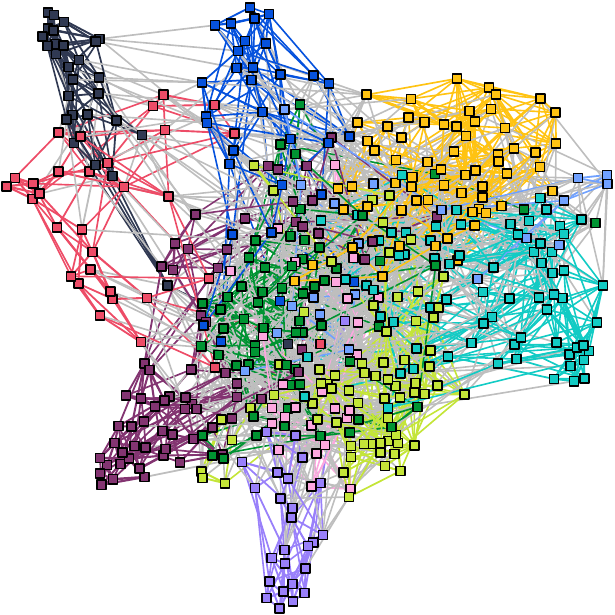}
        \caption{NGL}
    \end{subfigure}%
     \begin{subfigure}[t]{0.3\textwidth}
        \centering
        \includegraphics[scale=.36]{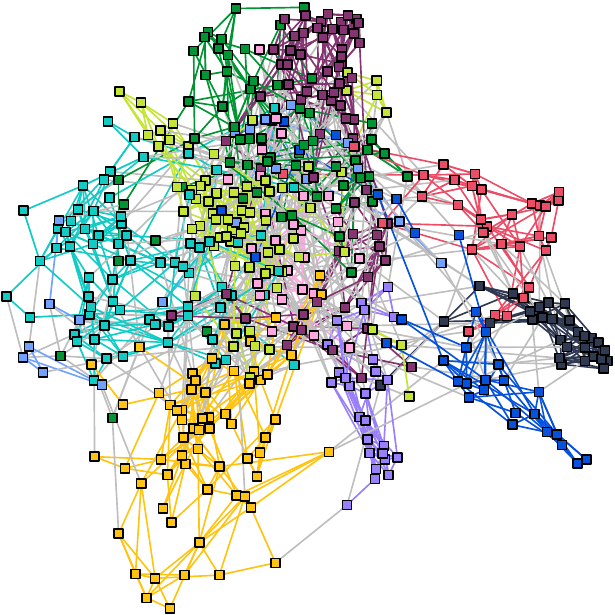}
        \caption{Proposed}
    \end{subfigure}%
    \caption{ Stock graphs learned via (a) GLE-ADMM, (b) NGL, and (c) Proposed method. The \textit{modularity} values for GLE-ADMM, NGL, and our proposed method are 0.4253, 0.4956, and 0.6207, respectively.}\label{fig:stock}
\end{figure*}

To evaluate the performance of edge recovery, we utilize the F-score ($\mathsf{FS}$) metric, which is defined as:
\begin{equation}
\mathsf{FS}=\frac{2\mathsf{tp}}{2\mathsf{tp}+\mathsf{fp}+\mathsf{fn}},
\end{equation}
where $\mathsf{tp}$, $\mathsf{fp}$, and $\mathsf{fn}$ represents true positives, false positives, and false negatives, respectively. The F-score values range between $0$ and $1$, with a score of $1$ indicating perfect identification of all connections and non-connections in the ground-truth graphs. The curves displayed in Figure \ref{fig:er-fs} are the average results of 10 Monte Carlo simulations.

Figure~\ref{fig:er-fs} demonstrates that our proposed method achieves an F-score of $1$, indicating the correct identification of all graph edges when the sample size is sufficiently large. Moreover, our method generally requires fewer samples than NGL and GLE-ADMM to attain a specific F-score value.

\subsection{Real-world Data}

We perform numerical experiments on a financial time-series dataset to evaluate the effectiveness of edge recovery. The dataset comprises 485 stocks that constitute the S\&P 500 index, with data spanning from January 5, 2016, to July 1, 2020. This results in 1142 observations per stock.  We construct the log-returns data matrix as:
\begin{equation*}
X_{ij} = \log \pi_{i,j} - \log \pi_{i-1, j},
\end{equation*}
where $\pi_{i,j}$ denotes the closing price of the $j$-th stock on the $i$-th day. The stocks are categorized into 11 sectors by the global industry classification standard (GICS) system.

To measure the performance of edge recovery for financial time-series data, we employ the \textit{modularity} metric \cite{newman2006modularity}. A stock graph with high \textit{modularity} exhibits dense connections among stocks belonging to the same sector and sparse connections between stocks from different sectors. Thus, a higher \textit{modularity} value suggests a more accurate representation of the actual stock network.

Figure \ref{fig:stock} presents the outstanding performance of our proposed method in edge recovery when compared to GLE-ADMM and NGL. This superiority is demonstrated by the fact that our method’s graph predominantly features connections between vertices within the same sector, while maintaining a minimal number of connections (represented by gray-colored edges) between vertices across different sectors. With \textit{modularity} values of 0.4253, 0.4956, and 0.6207 for GLE-ADMM, NGL, and our method, respectively, our approach showcases improved interpretability and more precise edge recovery than the competing techniques.

\section{Conclusions}\label{conclusions}

In this paper, we have proposed a novel formulation incorporating sparsity and Laplacian constraints for inferring network topology. Our approach addresses the limitations of the $\ell_1$-norm regularization in Laplacian constrained Gaussian graphical models while effectively promoting sparsity. We have introduced an efficient gradient projection algorithm to solve the resulting problem. The efficacy of our method has been demonstrated through numerical experiments on both synthetic and financial time-series datasets.

\bibliographystyle{IEEEtran}

\begin{thebibliography}{10}
\providecommand{\url}[1]{#1}
\csname url@samestyle\endcsname
\providecommand{\newblock}{\relax}
\providecommand{\bibinfo}[2]{#2}
\providecommand{\BIBentrySTDinterwordspacing}{\spaceskip=0pt\relax}
\providecommand{\BIBentryALTinterwordstretchfactor}{4}
\providecommand{\BIBentryALTinterwordspacing}{\spaceskip=\fontdimen2\font plus
\BIBentryALTinterwordstretchfactor\fontdimen3\font minus
  \fontdimen4\font\relax}
\providecommand{\BIBforeignlanguage}[2]{{%
\expandafter\ifx\csname l@#1\endcsname\relax
\typeout{** WARNING: IEEEtran.bst: No hyphenation pattern has been}%
\typeout{** loaded for the language `#1'. Using the pattern for}%
\typeout{** the default language instead.}%
\else
\language=\csname l@#1\endcsname
\fi
#2}}
\providecommand{\BIBdecl}{\relax}
\BIBdecl

\bibitem{ying2020nonconvex}
J.~Ying, J.~V. d.~M. Cardoso, and D.~P. Palomar, ``Nonconvex sparse graph
  learning under {Laplacian} constrained graphical model,'' in \emph{Advances
  in Neural Information Processing Systems}, vol.~33, 2020, pp. 7101--7113.

\bibitem{shuman2013emerging}
D.~I. Shuman, S.~K. Narang, P.~Frossard, A.~Ortega, and P.~Vandergheynst, ``The
  emerging field of signal processing on graphs: {Extending} high-dimensional
  data analysis to networks and other irregular domains,'' \emph{IEEE Signal
  Processing Magazine}, vol.~30, no.~3, pp. 83--98, 2013.

\bibitem{ortega2018graph}
A.~Ortega, P.~Frossard, J.~Kova{\v{c}}evi{\'c}, J.~M. Moura, and
  P.~Vandergheynst, ``Graph signal processing: {Overview}, challenges, and
  applications,'' \emph{Proceedings of the IEEE}, vol. 106, no.~5, pp.
  808--828, 2018.

\bibitem{dong2019learning}
X.~Dong, D.~Thanou, M.~Rabbat, and P.~Frossard, ``Learning graphs from data:
  {A} signal representation perspective,'' \emph{IEEE Signal Processing
  Magazine}, vol.~36, no.~3, pp. 44--63, 2019.

\bibitem{mateos2019connecting}
G.~Mateos, S.~Segarra, A.~G. Marques, and A.~Ribeiro, ``Connecting the dots:
  {Identifying} network structure via graph signal processing,'' \emph{IEEE
  Signal Processing Magazine}, vol.~36, no.~3, pp. 16--43, 2019.

\bibitem{9244648}
A.~G. Marques, S.~Segarra, and G.~Mateos, ``Signal processing on directed
  graphs: {The} role of edge directionality when processing and learning from
  network data,'' \emph{IEEE Signal Processing Magazine}, vol.~37, no.~6, pp.
  99--116, 2020.

\bibitem{leus2023graph}
G.~Leus, A.~G. Marques, J.~M. Moura, A.~Ortega, and D.~I. Shuman, ``Graph
  signal processing: {History}, development, impact, and outlook,'' \emph{IEEE
  Signal Processing Magazine}, vol.~40, no.~4, pp. 49--60, 2023.

\bibitem{ying2020does}
J.~Ying, J.~V. d.~M. Cardoso, and D.~P. Palomar, ``Does the $\ell_1$-norm learn
  a sparse graph under {Laplacian} constrained graphical models?'' \emph{arXiv
  preprint arXiv:2006.14925}, 2020.

\bibitem{kumar2019unified}
S.~Kumar, J.~Ying, J.~V. d.~M. Cardoso, and D.~P. Palomar, ``A unified
  framework for structured graph learning via spectral constraints,''
  \emph{Journal of Machine Learning Research}, vol.~21, no.~22, pp. 1--60,
  2020.

\bibitem{dong2016learning}
X.~Dong, D.~Thanou, P.~Frossard, and P.~Vandergheynst, ``Learning {Laplacian}
  matrix in smooth graph signal representations,'' \emph{IEEE Transactions on
  Signal Processing}, vol.~64, no.~23, pp. 6160--6173, 2016.

\bibitem{banerjee2008model}
O.~Banerjee, L.~E. Ghaoui, and A.~d'Aspremont, ``Model selection through sparse
  maximum likelihood estimation for multivariate {Gaussian} or binary data,''
  \emph{Journal of Machine Learning Research}, vol.~9, pp. 485--516, 2008.

\bibitem{d2008first}
A.~d'Aspremont, O.~Banerjee, and L.~El~Ghaoui, ``First-order methods for sparse
  covariance selection,'' \emph{SIAM Journal on Matrix Analysis and
  Applications}, vol.~30, no.~1, pp. 56--66, 2008.

\bibitem{ying2021minimax}
J.~Ying, J.~V. d.~M. Cardoso, and D.~P. Palomar, ``Minimax estimation of
  {Laplacian} constrained precision matrices,'' in \emph{International
  Conference on Artificial Intelligence and Statistics}, 2021, pp. 3736--3744.

\bibitem{egilmez2017graph}
H.~E. Egilmez, E.~Pavez, and A.~Ortega, ``Graph learning from data under
  {Laplacian} and structural constrints,'' \emph{IEEE Journal of Selected
  Topics in Signal Processing}, vol.~11, no.~6, pp. 825--841, 2017.

\bibitem{zhao2019optimization}
L.~Zhao, Y.~Wang, S.~Kumar, and D.~P. Palomar, ``Optimization algorithms for
  graph {Laplacian} estimation via {ADMM} and {MM},'' \emph{IEEE Transactions
  on Signal Processing}, vol.~67, no.~16, pp. 4231--4244, 2019.

\bibitem{segarra2017network}
S.~Segarra, A.~G. Marques, G.~Mateos, and A.~Ribeiro, ``Network topology
  inference from spectral templates,'' \emph{IEEE Transactions on Signal and
  Information Processing over Networks}, vol.~3, no.~3, pp. 467--483, 2017.

\bibitem{cardoso2021graphical}
J.~V. D.~M. Cardoso, J.~Ying, and D.~P. Palomar, ``Graphical models in
  heavy-tailed markets,'' in \emph{Advances in Neural Information Processing
  Systems}, vol.~34, 2021, pp. 19\,989--20\,001.

\bibitem{cardoso2022learning}
------, ``Learning bipartite graphs: Heavy tails and multiple components,'' in
  \emph{Advances in Neural Information Processing Systems}, vol.~35, 2022, pp.
  14\,044--14\,057.

\bibitem{kumar2019bipartite}
S.~Kumar, J.~Ying, J.~V. d.~M. Cardoso, and D.~P. Palomar, ``Bipartite
  structured {Gaussian} graphical modeling via adjacency spectral priors,'' in
  \emph{2019 53rd Asilomar Conference on Signals, Systems, and Computers},
  2019, pp. 322--326.

\bibitem{cardoso2024nonconvex}
J.~V. d.~M. Cardoso, J.~Ying, and D.~P. Palomar, ``Nonconvex graph learning:
  sparsity, heavy tails, and clustering,'' in \emph{Signal Processing and
  Machine Learning Theory}.\hskip 1em plus 0.5em minus 0.4em\relax Elsevier,
  2024, pp. 1049--1072.

\bibitem{yuan2023joint}
Y.~Yuan, D.~W. Soh, K.~Guo, Z.~Xiong, and T.~Q.~S. Quek, ``Joint network
  topology inference via structural fusion regularization,'' \emph{IEEE
  Transactions on Knowledge and Data Engineering}, pp. 1--18, 2023.

\bibitem{fallat2017total}
S.~Fallat, S.~Lauritzen, K.~Sadeghi, C.~Uhler, N.~Wermuth, and P.~Zwiernik,
  ``Total positivity in {Markov} structures,'' \emph{The Annals of Statistics},
  vol.~45, no.~3, pp. 1152--1184, 2017.

\bibitem{lauritzen2019maximum}
S.~Lauritzen, C.~Uhler, and P.~Zwiernik, ``Maximum likelihood estimation in
  {Gaussian} models under total positivity,'' \emph{The Annals of Statistics},
  vol.~47, no.~4, pp. 1835--1863, 2019.

\bibitem{rossell2021dependence}
D.~Rossell and P.~Zwiernik, ``Dependence in elliptical partial correlation
  graphs,'' \emph{Electronic Journal of Statistics}, vol.~15, no.~2, pp.
  4236--4263, 2021.

\bibitem{slawski2015estimation}
M.~Slawski and M.~Hein, ``Estimation of positive definite {M-matrices} and
  structure learning for attractive {Gaussian} {Markov} random fields,''
  \emph{Linear Algebra and its Applications}, vol. 473, pp. 145--179, 2015.

\bibitem{ying2021fast}
J.~Ying, J.~V. d.~M. Cardoso, and D.~P. Palomar, ``A fast algorithm for graph
  learning under attractive {Gaussian} {Markov} random fields,'' in \emph{2021
  55th Asilomar Conference on Signals, Systems, and Computers}, 2021, pp.
  1520--1524.

\bibitem{pavez2016generalized}
E.~Pavez and A.~Ortega, ``Generalized {Laplacian} precision matrix estimation
  for graph signal processing,'' in \emph{2016 IEEE International Conference on
  Acoustics, Speech and Signal Processing}, 2016, pp. 6350--6354.

\bibitem{deng2020fast}
Z.~Deng and A.~M.-C. So, ``A fast proximal point algorithm for generalized
  graph {Laplacian} learning,'' in \emph{IEEE International Conference on
  Acoustics, Speech and Signal Processing (ICASSP)}, 2020, pp. 5425--5429.

\bibitem{cai2021fast}
J.-F. Cai, J.~V. D.~M. Cardoso, D.~P. Palomar, and J.~Ying, ``Fast projected
  {Newton-like} method for precision matrix estimation under total
  positivity,'' \emph{arXiv preprint arXiv:2112.01939}, 2021.

\bibitem{ying2023adaptive}
J.~Ying, J.~V. D.~M. Cardoso, and D.~P. Palomar, ``Adaptive estimation of
  graphical models under total positivity,'' in \emph{International Conference
  on Machine Learning}, 2023, pp. 40\,054--40\,074.

\bibitem{truell2021maximum}
M.~Truell, J.-C. H{\"u}tter, C.~Squires, P.~Zwiernik, and C.~Uhler, ``Maximum
  likelihood estimation for brownian motion tree models based on one sample,''
  \emph{arXiv preprint arXiv:2112.00816}, 2021.

\bibitem{chung1997spectral}
F.~R. Chung, \emph{Spectral graph theory}.\hskip 1em plus 0.5em minus
  0.4em\relax American Mathematical Soc., 1997, no.~92.

\bibitem{holbrook2018differentiating}
A.~Holbrook, ``Differentiating the pseudo determinant,'' \emph{Linear Algebra
  and its Applications}, vol. 548, pp. 293 -- 304, 2018.

\bibitem{ying2017hankel}
J.~Ying, H.~Lu, Q.~Wei, J.-F. Cai, D.~Guo, J.~Wu, Z.~Chen, and X.~Qu, ``Hankel
  matrix nuclear norm regularized tensor completion for $ n $-dimensional
  exponential signals,'' \emph{IEEE Transactions on Signal Processing},
  vol.~65, no.~14, pp. 3702--3717, 2017.

\bibitem{ying2018vandermonde}
J.~Ying, J.-F. Cai, D.~Guo, G.~Tang, Z.~Chen, and X.~Qu, ``Vandermonde
  factorization of {Hankel} matrix for complex exponential signal
  recovery—application in fast {NMR} spectroscopy,'' \emph{IEEE Transactions
  on Signal Processing}, vol.~66, no.~21, pp. 5520--5533, 2018.

\bibitem{cai2016robust}
J.-F. Cai, X.~Qu, W.~Xu, and G.-B. Ye, ``Robust recovery of complex exponential
  signals from random {Gaussian} projections via low rank {Hankel} matrix
  reconstruction,'' \emph{Applied and Computational Harmonic Analysis},
  vol.~41, no.~2, pp. 470--490, 2016.

\bibitem{newman2006modularity}
M.~E. Newman, ``Modularity and community structure in networks,''
  \emph{Proceedings of the National Academy of Sciences}, vol. 103, no.~23, pp.
  8577--8582, 2006.

\end{thebibliography}

\end{document}